\documentclass[A4paper, 11 pt, conference]{ieeeconf}
\IEEEoverridecommandlockouts       
\usepackage{geometry}
\usepackage{threeparttable}
\usepackage{textcomp}
\usepackage[noadjust]{cite}
\usepackage[utf8]{inputenc}
\usepackage[T1]{fontenc} 
\usepackage{amsmath}
\usepackage[cmintegrals]{newtxmath}
\usepackage{bm}
\usepackage{array}
\usepackage{url}
\usepackage[utf8]{inputenc}
\usepackage{graphicx}
\usepackage[T1]{fontenc}
\usepackage[utf8]{inputenc}
\usepackage{babel}
\usepackage[font=small,labelfont=bf,tableposition=top]{caption}
\usepackage{booktabs}
\usepackage{graphics} 
\usepackage{epsfig} 
\usepackage{mathptmx} 
\usepackage{times} 
\usepackage{amsmath} 
\usepackage{amssymb}  
\usepackage{subfigure}
\usepackage{algorithm} 
\usepackage{algorithmic} 
\usepackage{multirow} 

\usepackage{color,soul}

\begin{document}

\title{\bf Towards a Multispectral RGB-IR-UV-D Vision System --- Seeing the Invisible in 3D}
\author{Tanhao Zhang, Luyin Hu, Lu Li, and David Navarro-Alarcon%
\thanks{This work is supported in part by the Research Grants Council of Hong Kong under grant 15212721, in part by the Jiangsu Industrial Technology Research Institute Collaborative Research Program Scheme under grant ZG9V, and in part by the Chinese National Engineering Research Centre for Steel Construction Hong Kong Branch under grant BBV8.}
\thanks{Tanhao Zhang, Luyin Hu and David Navarro-Alarcon are with The Hong Kong Polytechnic University, Department of Mechanical Engineering, Hung Hom, Kowloon, Hong Kong. (Corresponding author's e-mail: dna@ieee.org)}
\thanks{Li Lu is with the Changzhou Institute of Advanced Technology, Changzhou, Jiangsu Province, China.}
}
\date{}

\maketitle 
\thispagestyle{empty}

\begin{abstract}
In this paper, we present the development of a sensing system with the capability to compute multispectral point clouds in real-time.
The proposed multi-eye sensor system effectively registers information from the visible, (long-wave) infrared, and ultraviolet spectrum to its depth sensing frame, thus enabling to measure a wider range of surface features that are otherwise hidden to the naked eye.
For that, we designed a new cross-calibration apparatus that produces consistent features which can be sensed by each of the cameras, therefore, acting as a multispectral ``chessboard''.
The performance of the sensor is evaluated with two different cases of studies, where we show that the proposed system can detect ``hidden'' features of a 3D environment.
\end{abstract}

\section{INTRODUCTION}
The use of consumable three-dimensional (3D) vision systems has significantly exploded in the past decade (particularly, with the ubiquitous RBG-D sensors). 
The availability of these affordable 3D cameras has triggered many developments in important areas such as artificial intelligence, autonomous systems, service robotics, autonomous driving \cite{auto_mobile}, to name a few cases. 
However, most existing depth sensors are only equipped with a visible spectrum camera, along which they form 3D models of the environment. In some special occasions, such as criminal investigation and energy auditing \cite{T-ICP} , visible spectrum is not enough. A 3D multispectral vision system that can perceive the electromagnetic wave period from ultraviolet (UV) to long-wave infrared (LWIR) is needed to detect some "hidden features". In this article, we are using a depth camera, a UV camera and a thermal (LWIR) camera to build 3D multispectral vision system and make a special tool to calibrate all these cameras.

Thermal camera is a useful tool to conduct measurement and analysis tasks. It measures the thermal information without contact and shows temperature in imaging way \cite{3D_thermal}. 3D camera can get both visible spectrum (red, green, blue) and depth information, which has been widely used in the robotics community \cite{3D_slam}. UV camera is used for quality inspection, body fluid trace detection, sulfur dioxide ($SO_2$) detection \cite{UV_SO2} and so on. In \cite{body fluid UV}, Springer \textit{et al.} proved that UV can be absorbed by body fluid because of amino acids in these secretions. In \cite{UV_Skin}, UV image is used for checking the skin issues of humans face because the pigment of skin absorbs more UV that show higher contrast image than visible spectrum image, thus easier and faster to detect some skin issues. 

In the past two decades, there have been many articles put forward their own way to build 3D models with thermal information in different ways. In \cite{3D-thermal}, Vidas \textit{et al.} fused thermal camera with a structure light 3D camera (Kinect v1) and proposed a new technique for calibration, which did not need artificial targets (checkerboard). However, they assumed that both two cameras had similar fields of view, which is really hard to achieve when we use low-cost thermal camera which is inapplicable for the wildly used low-cost thermal cameras since their field of views are relatively smaller than regular RGB cameras. In \cite{PST900}, Shivakumar \textit{et al.} reflectivity based calibration is used for getting the intrinsic and extrinsic parameters of the thermal and RGB-D camera. They use different materials that have different emissivities ($\epsilon$) of LWIR to make "black" and "white" parts of checkerboard that yield high contrast images in both the thermal camera and RGB camera. The calibration principle and processing is the same as RGB camera. However, this method needs a large and stable heat source to generate the "reflex", which is not convenient for setting the calibration environment. There are few articles related to building a 3D model with UV. Based on our current knowledge, this is the first time that a complete multi spectral RGB-IR-UV-D vision system is developed.

With the on-going COVID-19 pandemic, people are paying increasing attention to environmental hygiene and personal hygiene \cite{pandemic_thermal}. Our idea came out when we think if there is a way to build a 3D modeling system that can measure the temperature and find some place which is contaminated and needed to be clean. This system needs to be lightweight and can be attached to a mobile robot. The temperature and 3D information can be collected by a thermal camera and a depth camera. To detect the contamination, we decided to use UV camera because most of biohazardous pollutant contains amino acids and can be visualized by UV camera, and all information is in image format which means easier to be processed. To calibrate these cameras, we need to build a special calibration tool that can calibrate multi-spectrum and is flexible to use.

The original contribution of this work is twofold: (i) 
A new calibration tool is proposed to effectively cross-calibrate multispectral images; (ii) A 3D vision system is developed to automatically detect and register features in the LWIR and UV spectrum.

The rest of this article is organized as follows: Sec. II presents the design of the multispectral vision system; Sec III experimentally validates the performance of the system; Sec. IV gives final conclusions.

\section{METHODS} 

\begin{figure}[t]
    \centering
    \includegraphics[width =1 \columnwidth]{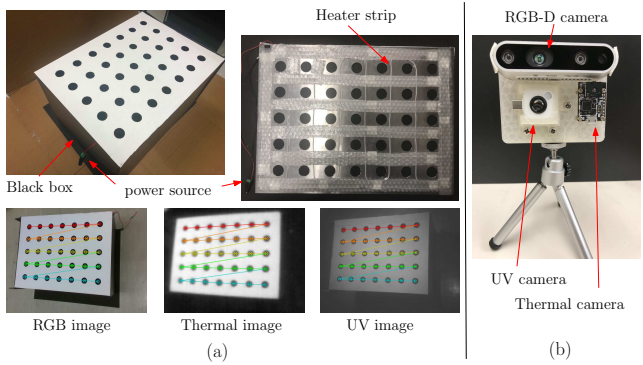}
    \caption{The calibration tool and result of calibration: (a): Top left - facade of calibration tool. Top right - revise side of calibration tool. (b): 3D Multispectral Vision System}
    \label{fig:calibration box}
\end{figure}

\subsection{A New Multispectral Calibration Tool}
Traditional calibration boards that use Zhang's calibration method \cite{Zhang's calibration} are checkerboards that use black and white blocks to make a high contrast in the image to detect feature points. However, for cameras which detect invisible light, i.e., UV (ultraviolet) cameras and LWIR (long wave infrared) cameras, are not suitable for this traditional calibration tool. We find that black blocks of the original calibration board can reflect a lot of UV although they can absorb most of visible light, thus feature points cannot be detected clearly by UV camera. And LWIR camera is sensitive to temperature. So, a newly designed calibration tool to calibration both visible light and invisible light is designed to calibration cameras which are sensitive to different spectrum.

The main part of the calibration tool is a rectangular aluminum plate with even-distributed circle holes on it and covered with white paper on the front. Heater strip is placed at the back of the plate. Heater strip will heat aluminum plate to make sufficient contrast for LWIR camera (thermal camera). To make sure that all cameras can be calibrated together, we make a black box that light can get in the box from holes on the aluminum board, but the main part of light is absorbed by reflecting many times in the black box, little get out from holes. Visible light and UV light are reflected by white paper (covered on the aluminum board) and absorbed by holes. As the Fig.\ref{fig:calibration box} shows, sufficient contrast can be detected by RGB camera, LWIR camera and UV camera. Once the features of the calibration tool (black circles) were detected, we mark the center of circle and use OpenCV's camera calibration function to get intrinsics of every camera and transformations matrices of each camera coordinates. 

\begin{figure}[t]
    \centering
    \includegraphics[width =1.0 \columnwidth]{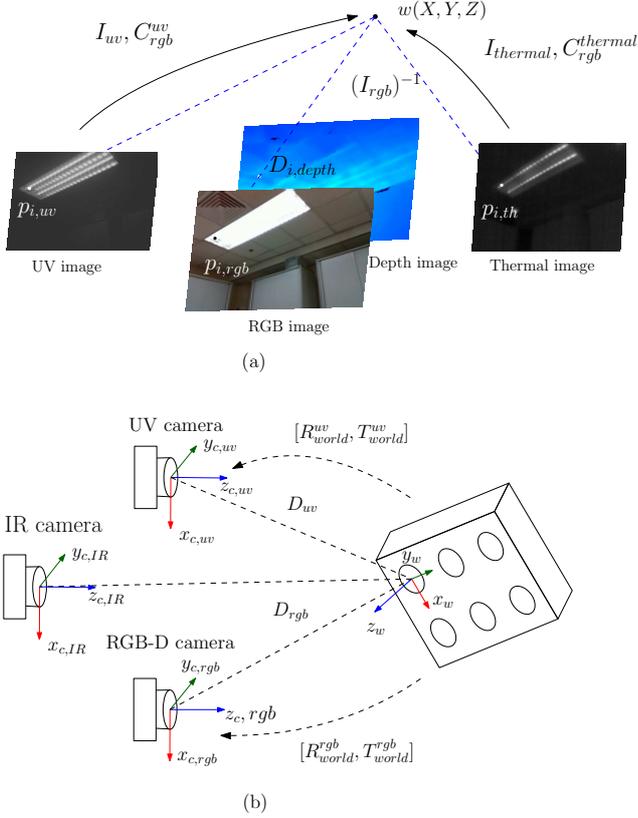}
    \caption{Alignment procedure of multispectral device: \textit{(a)} - Knowing intrinsics of each cameras, relative extrinsics and depth (distance), a point in RGB image can be projected to 3D world and then project to UV image and thermal image. \textit{(b)} - Getting the relative extrinsic of two cameras.}
    \label{fig:cams2world}
\end{figure}

\subsection{Alignment of 3D Multispectral Images}
3D multispectral visible system has three different kinds of camera, RGB-D camera, thermal camera and ultraviolet camera. Each camera can be simplified as a distorted pinhole camera model. The RGB-D camera is RealSense D435 depth camera which RGB and depth image have been aligned, thermal camera is a FLIR lepton 3.5, whereas the UV camera is a camera with ultraviolet sensitive CMOS and special lens which only pass ultraviolet which wavelength from 235nm to 395nm. Camera specifications are summarize in Table \ref{table:camera para}.
\begin{table}[t]
\caption{specifications of camera}\label{table:camera para}
\centering
\setlength{\tabcolsep}{0.7mm}
\begin{threeparttable}
\begin{tabular}{c|ccc}\toprule
\textbf{Sensor} &\textbf{Thermal camera} & \textbf{RGB-D camera} &\textbf{UV camera}\\ \\ \midrule
Model &FLIR lepton 3.5 &Realsense D435 &UVLOOK\\
    FOV & $57 deg.$ &$74 deg.$ &$55 deg.$\\
    Resolution & $160 \times120$ & $640 \times480$ & $640 \times 480$\\
    Frame rate & 8.7 FPS & 30 FPS & 30 FPS\\
    Wavelength & 9 - 14 µm & 400 - 800 nm & 235 - 395 nm \\ \bottomrule
\end{tabular}
\end{threeparttable}
\end{table}

We use Gaussian blur to resize images into same resolution. To align images of different cameras, we need to register each pixel from each image. RGB image is selected as reference image, which means other images need to be aligned with RGB image. Take RGB-UV camera alignment for example. $I_{rgb}$, $I_{uv}$ and $D_{rgb}$, $D_{uv}$ are the intrinsic and distortion coefficient of RGB camera and UV camera that obtained from intrinsic calibration respectively. Assuming $p_{rgb}=(u_r,v_r)$ and $p_{uv}=(u_u, v_u)$ are the pixel coordinates of RGB image and UV image. By using parameters above, we get image coordinates of undistorted RGB and UV images and their pixel coordinates $p_{u,rgb}$ and $p_{u,uv}$. 

To map each point of $p_{u,uv}$ to $p_{u,rgb}$, we first make each pixel include their own depth by using the depth image $D_{depth}$ acquired from the RealSense D435 to form a 3D array which can present the position of each pixel in the 3D world. And then projecting each point in 3D world of RGB camera to UV camera by using the relative camera coordinate $C^{uv}_{rgb}$, i.e the transformation matrix of RGB camera coordinate and UV camera coordinate. We use the homogeneous coordinate representation to denote 2D points and by $p_{u}=$[$u,v,1$]$^T$. For one point $i$, We can obtain a point position in 3D world coordinate $w_{i}$ by using the equation below:
\begin{equation}
\label{equ:world_point}
    w_{i} = ((I_{rgb})^{-1} \times p_{u,rgb}) \cdot D_{i,depth}
\end{equation}
Then we use the homogeneous coordinate represent 3D point $w_{i}=$[$X,Y,Z,1$]$^T$ and project $w_{i}$ to the UV image using the relative camera coordinate $C^{uv}_{rgb}$ to get $p_{i,u,uv}$:
\begin{equation}
\label{equ:UV_point_1}
    p_{i,u,uv} = I_{uv} \times (C^{uv}_{rgb} \times w_{i})
\end{equation}
Where vector $p_{i,u,uv}$ presents as homogeneous coordinate form. $C^{uv}_{rgb}$ include rotation and translation matrix of RGB camera and UV camera,$R_{rgb}^{uv}$ and $T_{rgb}^{uv}$, which present coordinate transformation of cameras.
\begin{equation}
    \label{equ:trans_matrix}
    C^{uv}_{rgb}=
    \left[ 
\begin{matrix}
R_{rgb}^{uv} & T_{rgb}^{uv} \\
0 & 1
\end{matrix}
\right ]
\end{equation}
To obtain $C^{uv}_{rgb}$, assuming $C_{uv}$, $C_{rgb}$ and $W_{b}$ are the UV camera coordinate, RGB camera coordinate and world coordinate respectively, and the calibration tool coordinate is regarded as the world coordinate, as the Fig.\ref{fig:cams2world} shows. RGB camera coordinate and world coordinate have relationship below:
\begin{equation}
    \label{equ:rcam2world}
    W_{b} =
    \left[
    \begin{matrix}
    R_{world}^{rgb} & T_{world}^{rgb} \\
    0 & 1
    \end{matrix}
    \right]
    ^{-1} C_{rgb}
\end{equation}
For the same reason, the relationship between UV camera coordinate and world coordinate is shown below:
\begin{equation}
    \label{equ:uvcam2world}
    W_b = 
    \left[
    \begin{matrix}
    R_{world}^{uv} & T_{world}^{uv} \\
    0 & 1
    \end{matrix}
    \right]
    ^{-1} C_{uv}
\end{equation}
Because RGB-D camera and UV camera are close to each other, the distances of camera coordinates can be regard as the same, we can get $C^{uv}_{rgb}$ as below:
\begin{equation}
    C_{rgb}^{uv} = 
    \left[
    \begin{matrix}
    R_{world}^{uv} & T_{world}^{uv} \\
    0 & 1
    \end{matrix}
    \right]
    \left[
    \begin{matrix}
    R_{world}^{rgb} & T_{world}^{rgb} \\
    0 & 1
    \end{matrix}
    \right]
    ^{-1}
\end{equation}
We now have the mapping from of UV camera to RGB camera. And we can also get the mapping from thermal camera to RGB camera by using the same method and the calibration tool. Now, we have aligned both UV image and thermal image to RGB image.

\begin{figure}[t]
    \centering
    \includegraphics[width = 1 \columnwidth]{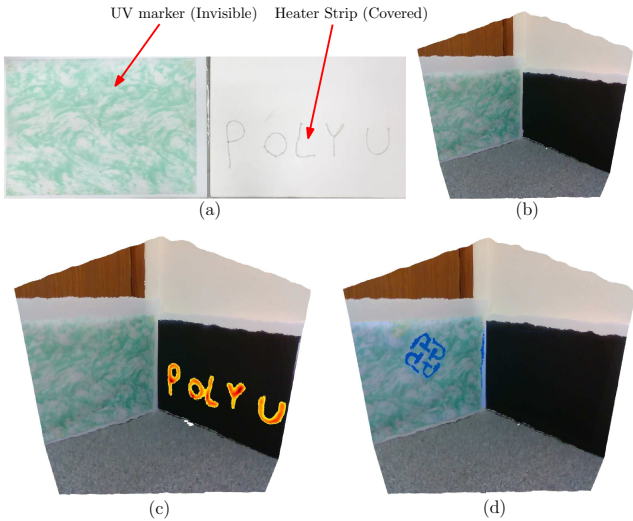}
    \caption{Testing board and alignment testing result: \textit{(a)} - Testing board contain two part, left part test UV-RGB alignment result,painted part is invisible for RGB camera. Right part test IR-RGB alignment result, we covered with a black paper on heater strip during the test. \textit{(b)(c)(d)} - Testing result, point clouds of RGB-D, RGB-Thermal-D (with thermal highlighted), RGB-UV-D (with UV highlighted).} 
    \label{fig:fusion_test}
\end{figure}

\subsection{Fusion of Multispectral Images}
Human beings are only able to see visible light. So thermal information is typically shown as visible colors with a colorbar to help the viewer estimate temperature of the area in thermal image \cite{colormap}. The same for UV images. There are three color channels (RGB, thermal, UV) and a depth channel needed to be shown. We can visualize depth and only one color channel as point cloud. Showing only one color channel each time will lost another two spectrum information, a problem remains when we want to show all spectrum together.

In this paper, we present a way to show all spectrum information in one color channel (RGB channel). In an ordinary way, visible spectrum contains most of the useful information. We can hardly confirm the texture of objects by using thermal image (as the Fig.\ref{fig:cams2world}(a) shows, we can not see the texture of the ceiling). So we only highlight some remarkable areas by using thermal image and UV image on RGB image to retain most of the visible spectrum information. Regarding this, we highlight the "hottest" image area of thermal image and the "most brightness" area of UV camera on RGB image by setting a threshold value $v_{thres,uv}$ and $v_{thers,thrmal}$ respectively, take UV image for example, the equation shows below:
\begin{equation}
    p_{u,rgb}=
    \begin{cases}
        p_{u,rgb} & p_{u,uv}\le v_{thres,uv}\\
        p_{u,uv} & p_{u,uv}>v_{thres,uv}
    \end{cases}
    \label{equ:fusion}
\end{equation}
We show the "hot area" as warm tones (red and yellow) and "UV area" as cool tones (blue and purple). Some image points do not have depth information or have invalid depth information, we call these points "bad points". For bad points, they cannot be mapped by using equation \ref{equ:world_point}, so we ignore these points which means they are still RGB points.

\begin{figure}[t]
    \centering
    \includegraphics[width = 1 \columnwidth]{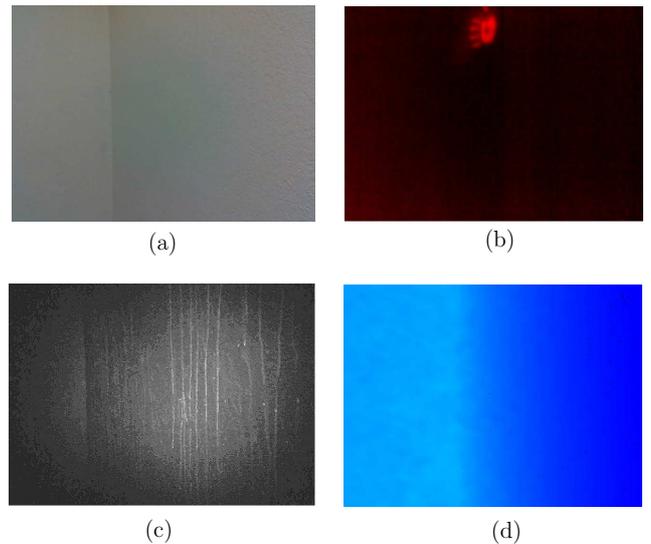}
    \caption{Different spectrum image of the wall: \textit{(a) (b) (c) (d)} - RGB, thermal, UV, depth image of the wall.} 
    \label{fig:fusion_result_2d}
\end{figure}
\begin{figure}[t]
    \centering
    \includegraphics[width = 1 \columnwidth]{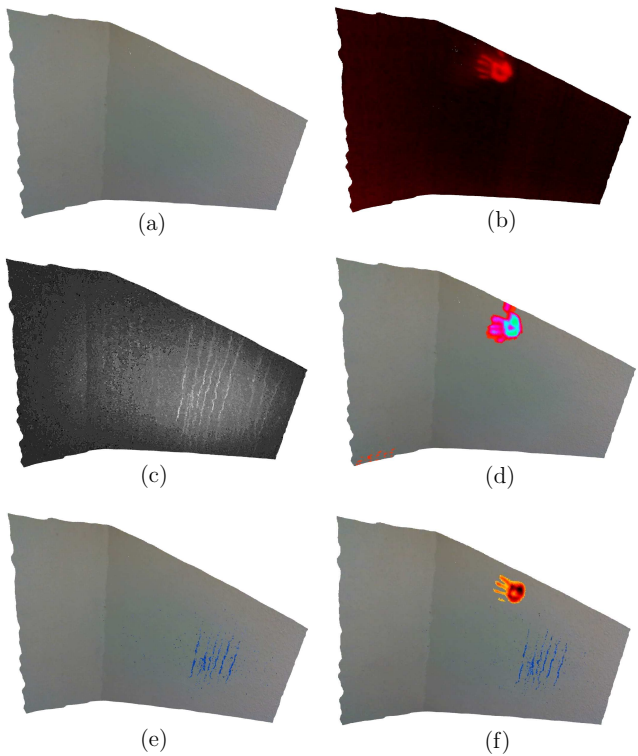}
    \caption{3D different spectrum and fusion of the wall:  \textit{(a) (b) (c)} - Point clouds of RGB-D, Thermal-D, UV-D. \textit{(d) (e) (f)} - Point clouds of RGB-thermal-D (with thermal highlighted), RGB-UV-D (with UV highlighted), RGB-thermal-UV-D (with both thermal and UV highlighted).} 
    \label{fig:fusion_result_3d}
\end{figure}

\section{RESULTS}
\subsection{Automatic 3D Image Alignment}
To test our alignment result, we build a special board that can both generate heat and reflect UV. This board contains two parts, one part for testing RGB-D-thermal alignment, the other part for testing RGB-D-UV alignment. The first part is a colored paper which is painted with titanium dioxide (which can reflect more UV then colored paper but not affect visible light) on some special region, to show the school badge on UV image. The other part is heater strip shows the word "PolyU" and covered with black paper so that the heater strip is invisible. We find few natural UV inside room, so a UV projector is needed when testing. We use Open3D \cite{open3d} to present the testing result, the testing result is shown as Fig \ref{fig:fusion_test}. Obviously, we have fused visible spectrum, UV, thermal and depth information in one point cloud. 

\subsection{Registration of ``Invisible'' Features}

Our 3D multispectral vision system has been tested in public toilet and yield good results. Some hidden features that cannot be observed by people's eyes are on display. We build multispectral (RGB, thermal, UV) point clouds of walls in the toilet. Our system can take down RGB, thermal, UV, and depth information together at the same time and form point clouds of RGB-D, Thermal-D, UV-D and RGB-Thermal-UV-D and each of them contain more than 306000 points. The mode of point cloud can be set during our program running. We highlight the thermal features in red and the UV features in blue. Due to limit UV inside room, the features of UV images are not easy to be separate, so we used Laplacian high-pass filter to sharpen the UV image, as Fig \ref{fig:fusion_result_2d} (c) shows. 

The Fig \ref{fig:fusion_result_3d} (a)-(f) show the point cloud of each spectrum and their fusion results. Fig \ref{fig:fusion_result_3d} (a) is the point cloud of a part of the wall, with few features shown on it. Fig \ref{fig:fusion_result_3d} (b) shows the remaining heat on the wall after someone touched and shows a hand shape. In Fig \ref{fig:fusion_result_3d} (c), we can observe some traces of dried body fluid running down the wall which cannot be seen under visible spectrum. Fig \ref{fig:fusion_result_3d} (d)-(f) illustrate the fusion results that use equation \ref{equ:fusion} for highlighting some interest areas of thermal image and UV image, which show some hidden features can be observed on visible spectrum point cloud. 

\section{CONCLUSION}
In this article, a novel calibration tool that can calibrate different spectrum together by using "Black Box Effect" is designed, and a 3D multispectral vision system that can observe the information of visible, thermal and UV spectrum is developed. The proposed calibration tool is easy to use and yields accurate result of intrinsic and extrinsic parameters of cameras. We then align images of each camera by using the obtained camera parameters. By highlighting the interest area of UV and thermal images on the RGB image, multispectral information can be emphasized in a point cloud. For future work, the developed multispectral vision system can be integrated with the mobile robot, which monitors people's body temperature by the thermal camera, detects the unseen containment by the UV camera, and navigates through the crowds by the depth camera.

\end{document}